\documentclass[letterpaper, 10 pt, conference]{ieeeconf}  

\IEEEoverridecommandlockouts                              

\usepackage{color}
\usepackage{amsmath}
\usepackage{amssymb}
\usepackage{cite}
\usepackage{graphicx}
\usepackage{subfigure}
\usepackage{balance}
\usepackage{multirow}
\usepackage{hyperref}
\usepackage{url}

\title{\LARGE \bf
Sparse Prototype Network for Explainable Pedestrian Behavior Prediction
}

\author{Yan Feng$^{*1}$, Alexander Carballo$^{2,3,4}$, and Kazuya Takeda$^{1,2, 4}$
\thanks{$^{1}$
Graduate School of Informatics, Nagoya University, Furo-cho, Chikusa-ku, Nagoya, Aichi 464-8601, Japan.
        }%
\thanks{$^{2}$
Institutes of Innovation for Future Society, Nagoya University, Furo-cho, Chikusa-ku, Nagoya, Aichi 464-8601, Japan.}%
\thanks{$^{3}$
Faculty of Engineering and Graduate School of Engineering, Gifu University, 1-1 Yanagido, Gifu City, 501-1193, Japan}%
\thanks{$^{4}$
Tier IV Inc., Nagoya University Open Innovation Center, 1-3, Mei-eki 1-chome, Nakamura-ward, Nagoya, 450-6610, Japan}%
}

\begin{document}

\maketitle
\thispagestyle{empty}
\pagestyle{empty}

\begin{abstract}

Predicting pedestrian behavior is challenging yet crucial for applications such as autonomous driving and smart city. Recent deep learning models have achieved remarkable performance in making accurate predictions, but they fail to provide explanations of their inner workings. One reason for this problem is the multi-modal inputs. To bridge this gap, we present Sparse Prototype Network (SPN), an explainable method designed to simultaneously predict a pedestrian's future action, trajectory, and pose. SPN leverages an intermediate prototype bottleneck layer to provide sample-based explanations for its predictions. The prototypes are modality-independent, meaning that they can correspond to any modality from the input. Therefore, SPN can extend to arbitrary combinations of modalities. Regularized by mono-semanticity and clustering constraints, the prototypes learn consistent and human-understandable features and achieve state-of-the-art performance on action, trajectory and pose prediction on TITAN and PIE. 
Finally, we propose a metric named Top-K Mono-semanticity Scale to quantitatively evaluate the explainability. Qualitative results show the positive correlation between sparsity and explainability. 
Code available at \url{https://github.com/Equinoxxxxx/SPN}.

\end{abstract}

\section{INTRODUCTION}

Predicting pedestrian behavior in complex environments is a critical task for autonomous systems, with applications ranging from self-driving vehicles to intelligent surveillance. While traditional models have focused on predicting either the future trajectory, action, or pose of pedestrians, there has been a growing demand for models that can simultaneously predict pedestrian behaviors in different types, including trajactory, action class and pose.\footnote{In this paper, the term ``behavior" refers to pedestrian trajectory, action class and pose unless otherwise specified.}

Recent efforts in applying deep learning methods to address the prediction of any one type of behavior are proven effective \cite{PCPA, pedgraphplus, JAAD, capformer}. However, only a few works discussed joint prediction of multiple types of behaviors \cite{PIE, girase2021loki}. Moreover, most of these works lack the mechanism of revealing their inner workings during inference, which causes additional testing costs when the model faces unseen scenarios, decreases the trustworthiness, and hinders developers and researchers from making further improvements. A major challenge that causes the problem is multi-modal inputs. Pedestrian behaviors can be inferred from various clues, including the historical trajectories, poses, contextual elements, etc. Although many of these clues can originate from visual inputs, the lack of annotated data and the high dimensionality of raw inputs make it hard for neural networks to learn salient features directly from images and videos. Therefore, disentangling different types of data and regarding them as multi-modalities became a common practice \cite{PCPA, capformer, PIE, pedgraphplus, STIP}. However, most explaining techniques are modality-specific \cite{ openaiprotocol} or architecture-specific \cite{CAM, gradcam, scorecam, rollout}, which limits the scalibility of the model. To bridge this gap, we present Sparse Prototype Network (SPN), a prototype-based framework designed to jointly predict multiple types of pedestrian behaviors, and provide explanations of its inferences based on the distance between the prototypes it learns and the input. The method is inspired by the idea that, multi-modalities derived from the same observation can be regarded as different fragments. By mapping the fragments into a joint latent space, a prototype vector can match any one of them, and is thus modality-independent, which enables the model to extend to arbitrary combinations of modalities. The modalities in the training data with the closest distance to a prototype are selected to represent that prototype. Figure \ref{fig1} briefly illustrates the inner working of SPN. 

\begin{figure}[t]
\centering
\includegraphics[width=0.48\textwidth]{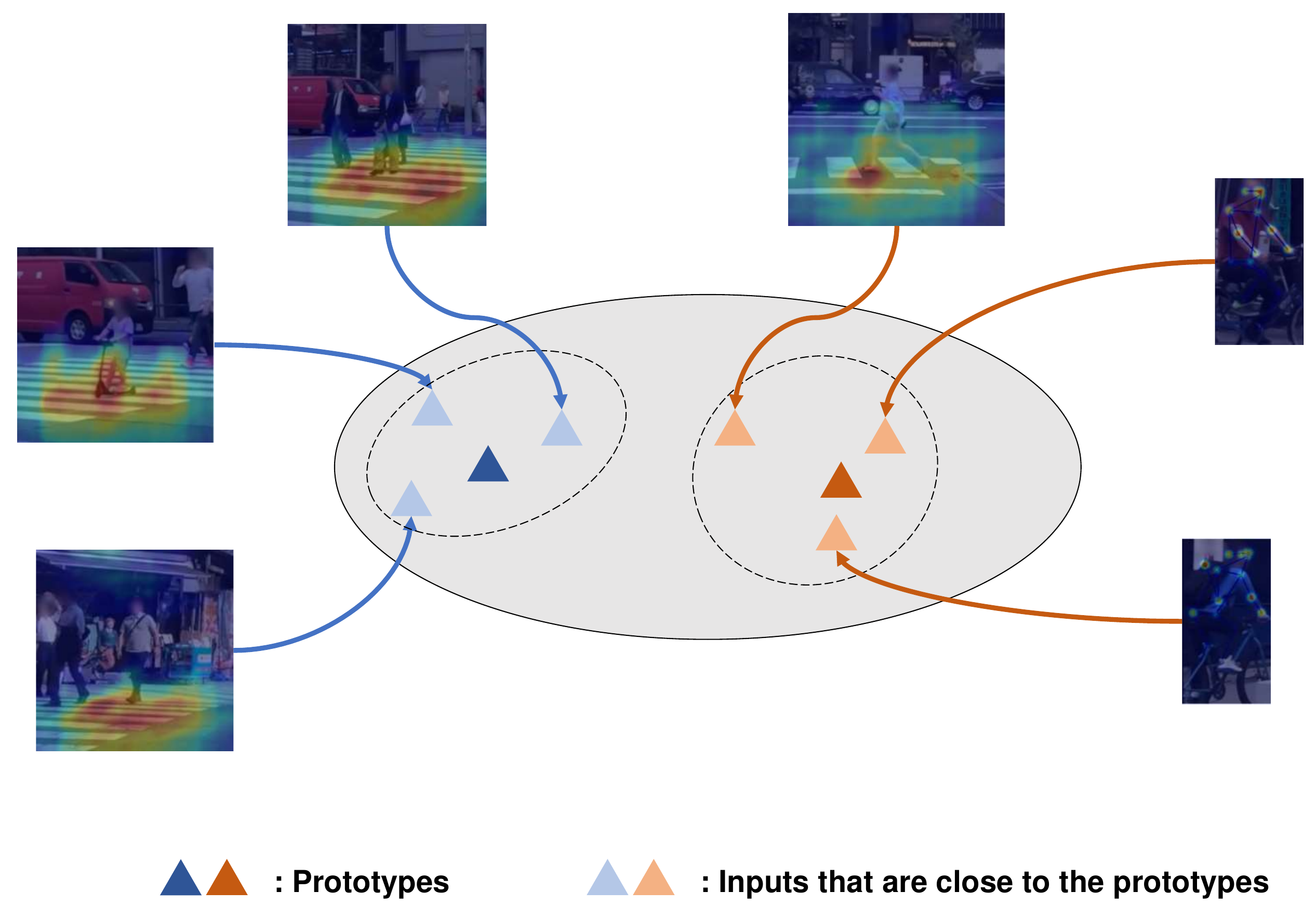}
\caption{Modality-independent prototypes. Multi-modal inputs are projected in a joint latent space, and the predictions are made based on their distance from the prototypes. Each prototype is represented by samples that are the closest to it.}
\label{figl}
\end{figure}

In the wider range of literature on explainable AI (XAI), one major challenge in developing explainable models is poly-semanticity, i.e. a unit of explaining (e.g. a neuron or a vector) activates on multiple semantics in the input. Since SPN uses prototype vectors to explain its inference, it also suffers from poly-semanticity. We argue that mono-semantic prototypes should activate sparsely on the whole dataset, since features that human understand are sparse \cite{monosemscale, encouragemono}. Motivated by this insight, we propose a sparsity loss term to encourage the prototypes to capture only sparse features, and thus to improve the mono-semanticity. Moreover, we propose a quantitative metric, named Top-K Mono-semanticity Scale, to evaluate the explainability of the prototypes without the influence of subjectivity brought by human annotators.

The contributions of this paper are as follows:

\begin{itemize}
\item We introduce SPN, a novel approach that simultaneously predicts pedestrian action, trajectory and pose while offering explainable predictions based on modality-independent prototypes. Therefore, SPN can be extended to arbitrary modalities. 

\item We propose a metric, the Top-K Mono-semanticity Scale, to quantitatively evaluate the explainability of the learned prototypes.

\item We apply a complementary combination of regularization losses to the prototypes. The sparsity term encourages the prototypes to capture mono-semantic features, while the clustering term provides the prototypes with the commonality between different modalities.

\item We validate SPN's performance on two popular pedestrian behavior prediction datasets, TITAN and PIE, and show that it achieves state-of-the-art performance while maintaining a high level of explainability.

\item We made the code available at \url{https://github.com/Equinoxxxxx/SPN}.
\end{itemize}

\section{RELATED WORKS}
\subsection{Pedestrian Behavior Prediction}
Traditional approaches for predicting pedestrian actions and trajectories relied on handcrafted features and physical models, such as Kalman filters and social force models \cite{socialforce}. These methods, while inherently explainable, often struggle to capture the complexity of real-world pedestrian behavior, especially in crowded or dynamic environments.

More recently, the field of pedestrian behavior prediction has witnessed significant advances brought by deep learning \cite{alahi2016sociallstm, gupta2018socialgan, peeking, PIE, JAAD, capformer}. State-of-the-art works primarily employed multi-modal inputs such as appearance, motion, past trajectory, etc \cite{PCPA, PIE, STIP, pedast, crossfeat, pedgraphplus, peeking}, and applied certain integration strategies, such as attention mechanisms, to make predictions. However, these models often lacked the ability to explain their predictions, limiting their applicability in safety-critical systems such as autonomous driving. 

\subsection{Prototype-based Models}
A large portion of efforts toward explainability in deep learning models have primarily focused on post-hoc explanation methods, where explanations are generated after the model has made its predictions \cite{CAM, gradcam, rollout, scorecam}. Recently, large language models (LLMs) have been applied to generate, text-based explanations, thanks to their outstanding ability of reasoning \cite{openaiprotocol, llmselfexplain}. However, these post-hoc explanations are limited in their fidelity. Moreover, they are often specific to a particular modality, such as visual data, and do not generalize well to multi-modal inputs. 

On the other hand, prototype-based methods \cite{protop,} offer an alternative path toward inherently explainable models. These methods learn prototypes—representative features—from the data, which are then used to make predictions. While most of these methods were applied to few-shot learning problem \cite{protoattribute, prototypicalnet, protounsupvideoseg}, some were designed to make explainable predictions in simple tasks and small-scale datasets \cite{protop, protopipnet}. The advantage of prototype-based approaches is that they provide sample-based explanations, i.e. using representative samples to explain the semantics of certain prototypes. However, these methods typically focus on single modalities and do not address the challenge of multi-modal explainability.

In order to bridge the above gaps for pedestrian behavior prediction, SPN leverages modality-independent prototypes, as well as a sparsity loss to promote mono-semanticity of the learned prototypes.

\subsection{Mono-semantic Explanation}
Sample-based explanation has recently become one of the major explaining techniques of XAI methods, including prototype-based methods \cite{conceptbottleneck, conceptwhitening, conceptframework, protop, protopipnet} and mechanistic methods \cite{openaiprotocol, mechainterhe2024dictionary}, where the most related samples are used to represent the explaining units such as neurons and prototypes. One recent challenge faced by these methods is poly-semanticity \cite{superposition}, meaning that the units are represented by unrelated samples. On the contrary, mono-semantic units refer to those that are related to relatively consistent features. Although works from the field of language models \cite{encouragemono, inhibitmonosem} used sample-wise metric to promote or inhibit the mono-semanticity of a single sample, there has not been a metric to evaluate the mono-semanticity of an explaining unit. Inspired by Mono-semantic Scale\cite{inhibitmonosem} where sparsity was used as a proxy of mono-semanticity, we propose Top-K Mono-semanticity Scale (Top-K MS) for short as a quantitative metric for explainability of prototypes.

\begin{figure*}[t]
\centering
\includegraphics[width=0.88\textwidth]{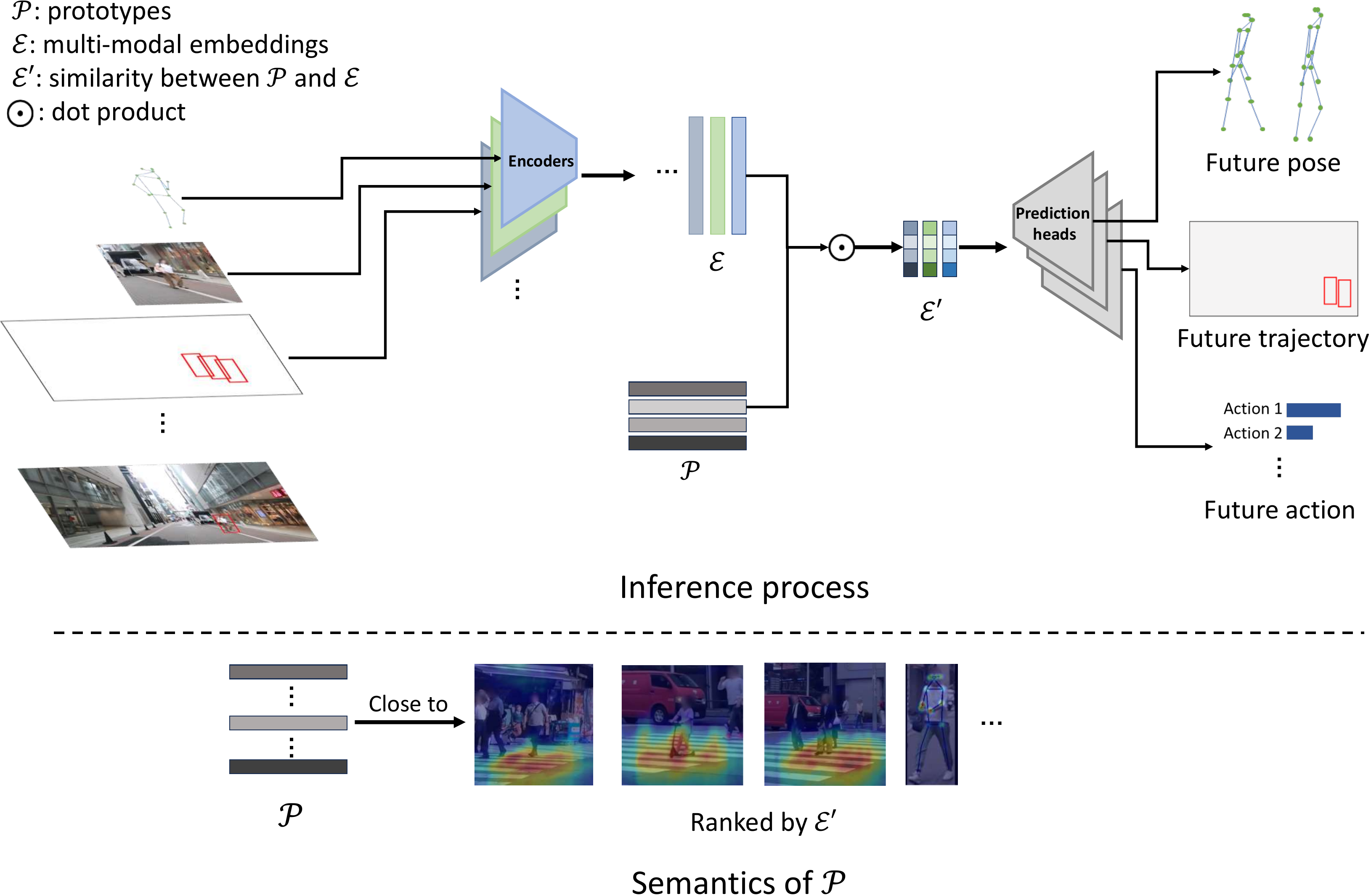}
\caption{Inference process of SPN and how the prototypes are explained.}
\label{fig2}
\end{figure*}

\section{Method}
In this section, we describe the architecture of the Sparse Prototype Network (SPN) and the explainability metric Top-K Mono-semanticity Scale (Top-K MS) used to evaluate the model. Our proposed method focuses on creating a predictive framework that not only achieves accurate pedestrian behavior prediction but also provides interpretable, human-understandable explanations for its decisions.

\subsection{Sparse Prototype Network}
The SPCM employs multi-modal input and predicts three types of data to describe pedestrian behavior: action, trajectory, and pose. To achieve this, the model is composed of three main modules: input encoding, prototype layer, and prediction heads.

\textbf{Input encoding.} The input encoding module processes each modality independently, transforming the raw inputs into compact, high-dimensional feature vectors. Let $\mathcal{X}=\left \{ \boldsymbol{x}_{m} \right \}^{M}_{m=1}$ be the inputs from $M$ modalities. The input encoding module consists of M encoders $\mathcal{F}=\left \{ {f}_{m} \right \}^{M}_{m=1}$. Each encoder ${f}_{m}$ corresponds to a modality $\boldsymbol{x}_{m}$, and is followed by an MLP layer to project the outputs into a joint latent space. The multi-mdoal embeddings are given by

\begin{eqnarray}
\begin{aligned}
\boldsymbol{e}_{m} = ReLU({W}_{e}{f}_{m}(\boldsymbol{x}_{m}))
\end{aligned}
\end{eqnarray}
where $\mathcal{E}=\left \{ \boldsymbol{e}_{m} \right \}^{M}_{m=1}$ represents the multi-modal embeddings, and ${W}_{e}$ is the weights of the MLPs.

In our experiments, we exploit up to 5 different modalities. Their formats are as follows:

1) Local context: a single image cropped by enlarged bounding box around the pedestrian. Following the practice in \cite{pedgraphplus}, we add semantic segmentation map as additional channels to the image. The encoder is in the same architecture as in \cite{pedgraphplus}, which is a multi-layer 2D convolutional block.

2) Past pose: a sequence of the past skeletons of the pedestrian in the COCO format. 

3) Past trajectory: a sequence of the past coordinates of bounding boxes of the target pedestrian in the image plane.

4) Ego motion: a sequence of acceleration of the ego vehicle.

5) Social relation: the relative location of other pedestrians and vehicles in the surroundings. Following the practice in \cite{peeking}, the format is given by

\begin{eqnarray}
\begin{aligned}
\mathcal {G}_{k}=[log(\frac{|x_b-x_k|}{w_b}), log(\frac{|y_b-y_k|}{h_b}), log(\frac{w_k}{w_b}), log(\frac{h_k}{h_b})]
\end{aligned}
\end{eqnarray}
where $\mathcal {G}_{k}$ is the relative location of the $k$-th neighbor.

\textbf{Prototype layer.} In this module, the multi-modal embeddings $\mathcal{E}$ is mapped to the similarity with as set of prototype vectors $\mathcal{P}=\left \{ \boldsymbol{p}_{n} \right \}^{N}_{n=1}$, where each $\boldsymbol{p}_{n}$ is in the same dimensions as $\boldsymbol{e}_{m}$. The matching process is given by

\begin{eqnarray}
\begin{aligned}
\mathcal{E}'=ReLU(\mathcal{P}^{\top}\mathcal{E})
\end{aligned}
\end{eqnarray}
where $\mathcal{E}'\in \mathbb{R}^{N\times M}$ is the matching results. The ReLU activation ensures that the similarity is non-negative and that the mapping is non-linear without losing interpretability \cite{superposition}. Although there is previous work \cite{protop} applying purely linear functions to ensure strictly interpretable architecture, non-linear activation in a single layer provides more possibilities of further improvements and investigations, such as more prototypes than dimensions \cite{superposition, monosemanticity}.

\textbf{Task-specific prediction heads.} Given the matching $\mathcal{E}'$ between prototypes and multi-modal embeddings, the predictions are made by a task-specific prediction head for each task. We frame trajectory prediction and pose prediction as generation task, and action prediction as classification task. For action prediction, we simply use a linear function to generate the action logits

\begin{eqnarray}
\begin{aligned}
\boldsymbol{\hat{y}}_{act}=softmax({W}_{act}\mathcal{E}')
\end{aligned}
\end{eqnarray}

For trajectory and pose prediction, we apply a model $g$ that supports conditioned generation and input $\mathcal{E}'$ as the condition that contains all past observations. The prediction is given by

\begin{eqnarray}
\begin{aligned}
\boldsymbol{\hat{y}}_{traj/pose}=g(\epsilon ;\mathcal{E}')
\end{aligned}
\end{eqnarray}
where $\epsilon$ is the Gaussian noise required for generation. More details about the implementation is introduced in Section IV-A.

\subsection{Loss Functions}
The model is optimized using a combination of two groups of losses:

\begin{itemize}
\item Task-specific Losses: For each prediction task (action, trajectory, and pose), we compute the respective loss. For action prediction, we use cross-entropy loss; for trajectory prediction and pose prediction, mean squared error (MSE).

\item Prototype Regularization Losses: Since the multi-modal inputs are fed to the prototype layer independently, no inter-modality interaction is maintained. Therefore, the underlying commonality between different modalities might be lost in the process. As compensation for this loss, a clustering term $\mathcal{L}_{cluster}$ is applied to encourage the prototypes to capture the commonality between different modalities. Specifically, we follow the practice in \cite{feng2024contrasting}, drawing the embeddings of the same sample together while pushing embeddings from different samples away from each other. The clustering loss is given by

\begin{eqnarray}
\begin{split}
&\mathcal{L}_{cluster} = 
\\&-\frac {1}{B M^2}\sum\limits_{i=1}^{B}\sum\limits_{m=1}^{M}\sum\limits_{n=1}^{M}log\frac{exp(\boldsymbol{e}_{m, i}\cdot \boldsymbol{e}_{n, i}/\tau)}{\sum\limits_{j=1}^{B}exp(\boldsymbol{e}_{m, i}\cdot\boldsymbol{e}_{n, j}/\tau)}
\end{split}
\end{eqnarray}
where $\boldsymbol{e}_{m, i}$ is the embedding for modality $m$ of the $i$-th sample, and $\tau$ is a temperature factor to normalize the dot product.

\end{itemize}
To avoid the multi-modal representations from collapsing to the same point, we also apply L1 loss $\mathcal{L}_{l1}$ as a counterpart to improve the sparsity degree of $\mathcal{E}'$, and meanwhile as a proxy of mono-semanticity.
The total loss is formulated as:

\begin{eqnarray}
\begin{aligned}
\mathcal{L} = {\lambda }_{cluster}{\mathcal{L}}_{cluster} + {\lambda }_{l1}{\mathcal{L}}_{l1} + {\mathcal{L}}_{task}
\end{aligned}
\end{eqnarray}

\subsection{Top-K Mono-semanticity Scale}
\begin{figure*}[t]
\centering
\includegraphics[width=0.78\textwidth]{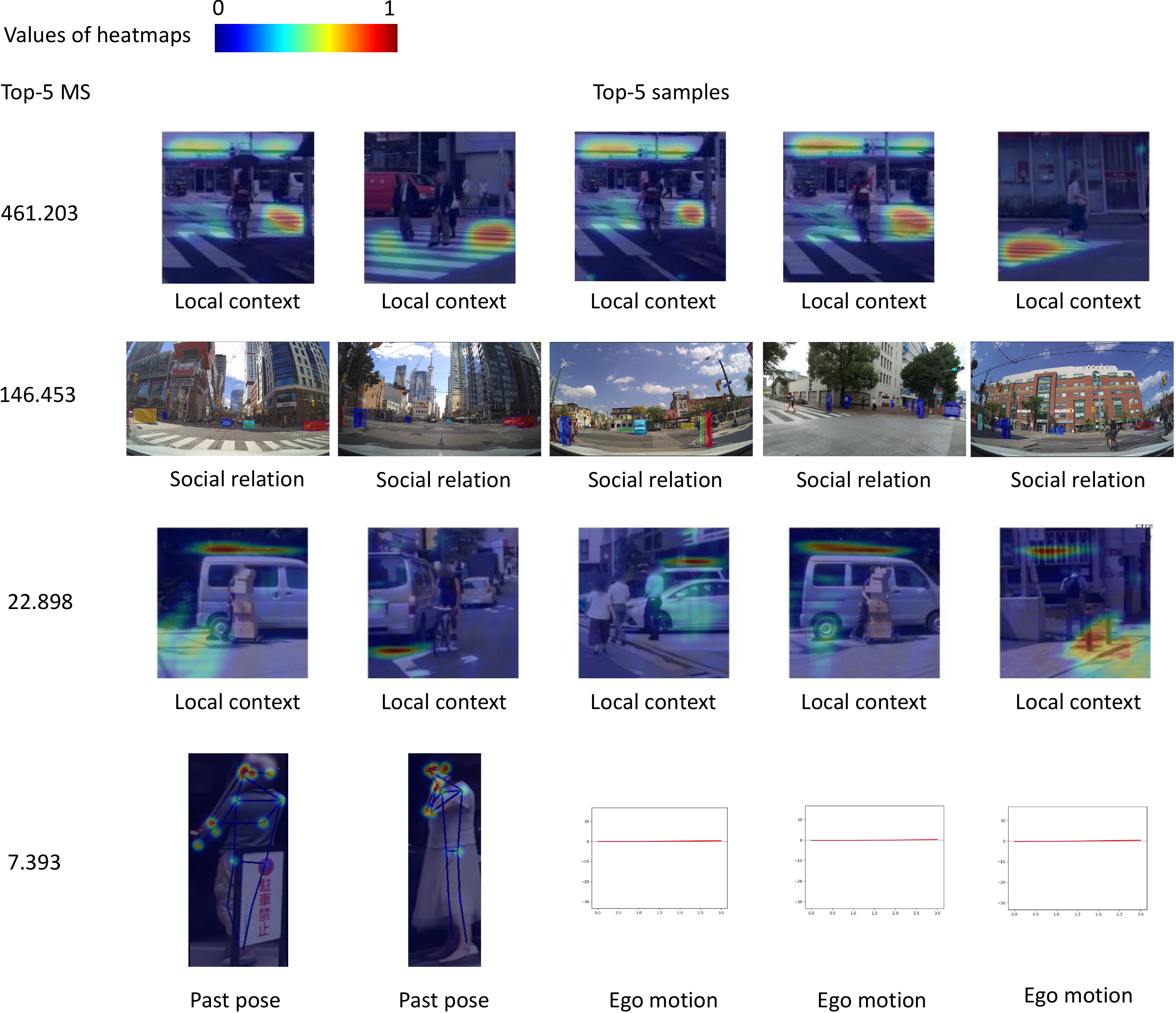}
\caption{Examples of prototypes with different Top-K MS. As the Top-K MS decreases, the semantics of the prototype become hard to recognize.}
\label{fig3}
\end{figure*}

Inspired by the Mono-semantic Scale \cite{inhibitmonosem} that uses sparsity as a proxy of mono-semanticity, we introduce the Top-K Mono-semanticity Scale to quantitatively evaluate the explainability of our model. This metric measures how well the top-K most activated are far from the average values. For each prototype, we rank the samples based on their activation strength and compute the mean relative variance of the top-K activations. Specifically, given the matching results between the prototypes and a batch of inputs of size B, the Top-K Mono-semanticity Scale (Top-K MS) is defined as

\begin{eqnarray}
\begin{aligned}
{\psi }_{n} = \frac {1}{K}\sum\limits_{k=1}^{K}\frac{\mathcal{E}'_{n,k}-\bar{\mathcal{E}'_{n}}}{S^{2}}
\end{aligned}
\end{eqnarray}
where
\begin{eqnarray}
\begin{aligned}
\bar{\mathcal{E}'_{n}}=\frac {1}{BM}\sum\limits_{i=1}^{BM}\mathcal{E}'_{n,i}, 
S^{2}=\frac {1}{BM-1}\sum\limits_{i=1}^{BM}(\mathcal{E}'_{n,i}-\bar{\mathcal{E}'_{n}})^2
\end{aligned}
\end{eqnarray}
and $\mathcal{E}'_{n,k}$ ranked in the descending order. A high ${\psi }_{n}$ value means that the n-th prototype has high activations on a limited number of samples, and the feature is thus sparse. Figure \ref{fig3} \footnote{See \url{https://github.com/Equinoxxxxx/SPN} for more visualization results.} illustrates prototypes with different degrees of Top-K MS and the samples with top activations. In the first row where the Top-K MS is high, the top samples are all in the same modality, and the highlighted regions focus on the crosswalks at a relatively far distance; in the second row, the prototype seems to focus on such a layout of agents where the target pedestrian is at the left edge of the view, and a surrounding vehicle or pedestrian on the opposite side is attended to; in the third row, the prototype focuses on a seemingly region of shadow; and in the last row, where the Top-K MS is the lowest, the prototype seems to capture both the head joints where the pedestrian is standing and a near-zero ego acceleration sequence. All in all, as the Top-5 MS values decrease, the semantics of the representative samples of a prototype tend to become less consistent and hard to understand.

\section{Experiments}
\subsection{Implementation Details}
The detailed settings of the multi-modal encoders we used in the experiments are as follows:

1) Local context. We use the backbone in \cite{pedgraphplus}, a multi-layer 2D convolutional block.

2) Past pose, past trajectory, ego motion and social relation.
For these modalities, we use a one-layer transformer encoder \cite{vaswani2017attention} with 8 heads and 64 dimensions. For modalities with more than one sequence dimension, we flatten these dimensions together with the temporal dimension before feeding the input to the backbone.

For the prediction heads of trajectory prediction and pose prediction, we use DePOSit \cite{deposit}, a diffusion-based generation model originally designed for pose regression. We input $\mathcal{E}'$ as a condition vector into DePOSit for pose and trajectory prediction.

We use Adam optimizer during training with $\beta _1=0.9$, $\beta _2=0.999$. We set ${\lambda }_{cluster}=0.001$ and ${\lambda }_{l1}=0.01$ out of a random search. We use 50 prototypes and 512 dimensions for each prototype as a default setting. The model is trained for 50 epochs with a batch size of 64.

\subsection{Datasets}
We evaluate SPN on PIE \cite{PIE} and TITAN \cite{TITAN}. TITAN contains 10 hours of 60 FPS driving videos in Tokyo and multiple independent action sets at 10 Hz frequency: communicative actions, transportive actions, complex contextual actions, simple contextual actions and atomic actions. PIE is a commonly used dataset for pedestrian action and trajectory prediction. It contains 6 hours of 30 FPS driving videos in Toronto with annotations of binary crossing action labels at 30 Hz frequency. Since PIE only contains action labels of pedestrian crossing the road, we use crossing prediction to evaluate the action prediction performance. 

\subsection{Evaluation Metrics}
We evaluate SPN using standard prediction metrics. For action prediction, we use accuracy and F1-score; for trajectory and pose prediction, we use mean squared error (MSE). Since TITAN and PIE have different image sizes, we normalize the trajectory and pose coordinates by their corresponding image size, and calculate the MSE on the normalized data.

\subsection{Behavior Prediction Performance}
We compare SPN on all three tasks with state-of-the-art methods, including SGNet \cite{sgnet}, next \cite{peeking}, PCPA \cite{PCPA}, Pedestrian Graph+ \cite{pedgraphplus} and DePOSit \cite{deposit}. Note that we concatenate TITAN and PIE together to evaluate the ability of the models with maximum data availability. And since SGNet, DePOSit and SPN are stochastic when predicting trajectory and pose, we choose the best result out of 5 runs as the final result.

\begin{table}[]
\caption{Behavior prediction results on TITAN\&PIE}
\label{tab1}
\scalebox{1.2}{
\begin{tabular}{lcccc}
\hline
\multirow{2}{*}{Models} & \multicolumn{4}{c}{TITAN\&PIE}                                 \\ \cline{2-5} 
                        & Acc           & F1            & Traj MSE      & Pose MSE       \\ \hline
SGNet                   & -             & -             & 1.2          & -              \\
SGNet-CVAE              &               &               & 1.0          &                \\
DePOSit                 & -             & -             & -             & 0.16          \\
Next                    & 0.72          & 0.57          & 0.11         & -              \\
PCPA                    & \textbf{0.78} & 0.45          & -             & -              \\
PedGraph+               & 0.77          & 0.51          &               &                \\
SPN(ours)               & \textbf{0.78} & \textbf{0.59} & \textbf{0.09} & \textbf{0.07} \\ \hline
\end{tabular}}
\end{table}

The results in Table \ref{tab1} show that SPN outperforms other models on most metrics. Note that PCPA and Pedestrian Graph+ have high accuracy but low F1, indicating that they are biased to the dominant class, since neither TITAN or PIE are balanced on the crossing action. For trajectory prediction and pose prediction, SPN exceeds other models, especially autoregressive models such as SGNet and DePOSit, to a considerable extent, indicating that the prototypes effectively encoded the rich information from multi-modal inputs and improved the prediction based on this information, despite the compactness of the prototype-based representations.

\begin{figure*}[t]
\centering
\includegraphics[width=0.80\textwidth]{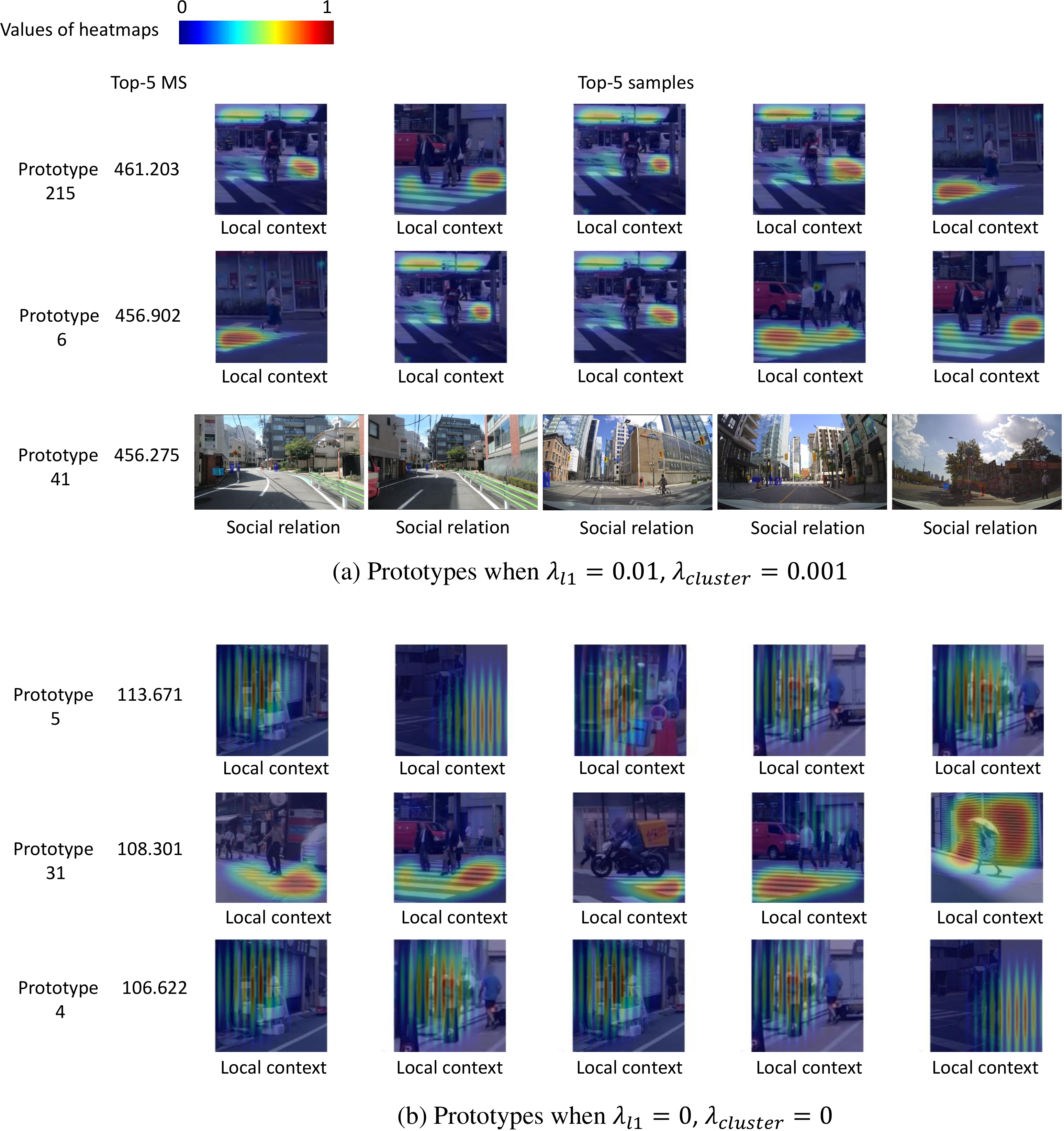}
\caption{Qualitative comparison between SPN with the regularization terms and SPN without them.}
\label{fig4}
\end{figure*}

\begin{table}[]
\caption{Ablation experiments on regularization terms}
\label{tab2}
\scalebox{0.9}{
\begin{tabular}{lccccc}
\hline
\multirow{2}{*}{Models}                                                                       & \multicolumn{5}{c}{TITAN\&PIE}                                                                                             \\ \cline{2-6} 
                                                                                              & Acc           & F1            & Traj MSE       & Pose MSE       & \begin{tabular}[c]{@{}c@{}}Mean \\ Top-5 MS\end{tabular} \\ \hline
\begin{tabular}[c]{@{}l@{}}${\lambda }_{cluster}=0$\\ ${\lambda }_{l1}=0$\end{tabular}        & \textbf{0.79} & 0.55          & 0.095          & 0.067          & 68.91                                                    \\
\begin{tabular}[c]{@{}l@{}}${\lambda }_{cluster}=0.001$\\ ${\lambda }_{l1}=0$\end{tabular}    & 0.49          & 0.47          & 0.110          & 0.069          & 34.83                                                    \\
\begin{tabular}[c]{@{}l@{}}${\lambda }_{cluster}=0$\\ ${\lambda }_{l1}=0.01$\end{tabular}     & 0.32          & 0.32          & 0.221          & 0.210          & 78.55                                                    \\
\begin{tabular}[c]{@{}l@{}}${\lambda }_{cluster}=0.001$\\ ${\lambda }_{l1}=0.01$\end{tabular} & \textbf{0.79} & \textbf{0.59} & \textbf{0.093} & \textbf{0.066} & \textbf{306.03}                                          \\ \hline
\end{tabular}
}
\end{table}

In Table \ref{tab2} we list the results of ablation experiments on the regularization terms by removing one or both of them. We also add the Top-5 MS value as a reference of the explainability. It can be seen from the second and third row that, either the clustering term or the L1 term alone would cause training collapse, meaning the model falls to a suboptimal. Only when both terms are present, the performance continue to improve. Beside the improvement on the prediction performance, the difference brought by the combination of two loss terms can also be seen in Figure \ref{fig4}, where prototypes with high Top-5 MS from the two cases are shown. While the upper part shows prototypes with relatively consistent and clear semantics, the lower part, where no regularization terms are applied, shows the tendency of focusing on patterns that are hard to understand, as well as lower Top-5 MS values. The comparison shows the effect of the regularization terms in not only improving prediction but also encouraging the prototypes to focus on more meaningful semantics.

\subsection{Evaluation of Partial Prototypes}
To further test the functionality of the prototypes with high sparsity, we conduct an experiment to evaluate the performance of SPN with only a small portion of prototypes functioning. Specifically, we select 5 out of 50 prototypes from a trained SPN, and set the rest prototypes to 0. The 5 prototypes are selected by two different criteria:

\begin{itemize}
    \item Top-5 MS. Prototypes with the highest Top-5 MS values are selected.
    \item L1 norm of the corresponding columns in $W_{act}$. Since $W_{act}$ transforms the prototype matching results $\mathcal{E}'$ into action logits, each column of $W_{act}$ corresponds to a certain prototype, where each element represents the relation between the prototype and one specific action class. By selecting columns of $W_{act}$ with the high L1 norm values, we are actually selecting prototypes with the highest importance to action prediction.
\end{itemize}

The results of the partial prototypes selected by the above two criteria are listed in Table \ref{tab3}. While the performances of trajectory and pose prediction are the same, it is interesting that prototypes selected by Top-5 MS have better performances. This could indicate that prototypes with high sparsity actually contains more effective action-related information than the prototypes with high importance in $W_{act}$, which are supposed to be the optimal sets for action prediction. Such a phenomenon can be evidence that sparse features could contain more meaningful semantics, even though not paid much attention by the model itself.

\begin{table}[]
\caption{Experiments on partial prototypes}
\label{tab3}
\scalebox{1.1}{
\begin{tabular}{lcccc}
\hline
\multirow{2}{*}{Selection metric} & \multicolumn{4}{c}{TITAN\&PIE}                                  \\ \cline{2-5} 
                                  & Acc           & F1            & Traj MSE       & Pose MSE       \\ \hline
Top-5 MS                          & \textbf{0.75} & \textbf{0.58} & \textbf{0.092} & \textbf{0.066} \\
Linear weights                    & 0.63          & 0.49          & \textbf{0.092} & 0.067          \\ \hline
\end{tabular}
}
\end{table}

\section{Limitations}
Despite the improvements in prediction performance and in the mono-semanticity of explanations, the above results also reveal limitations of SPN. First of all, mono-semanticity is only one aspect of general explainability, and applying Top-K MS and sparsity loss cannot totally diminish the subjectivity from the evaluation of explainability. Secondly, despite the fact that the classification task can be made completely interpretable by using linear functions as the prediction head, the transformation from prototypes to the generation results, i.e. trajectory and pose, still remains a black box. Although the prototypes can provide more transparent conditions than previous methods, more efforts are required to improve the explainability of the generation and regression process. Last but not least, although SPN shows progress with a relatively small number of prototypes, the potential still remains to be discovered to have more prototypes than dimensions such that the prototypes can learn further disentangled and fine-grained features from multi-modalities.

\section{CONCLUSIONS}
In this paper, we introduced the Sparse Prototype Network (SPN), an explainable architecture designed to address the challenge of explainable pedestrian behavior prediction. SPN surpasses existing models in three tasks, i.e. action prediction, trajectory prediction and pose prediction. By mapping multi-modal inputs to modality-independent prototypes, SPN can trace its own decisions to human-understandable concepts. To compensate for the lost commonality between modalities, we also apply a complementary combination of regularization terms, preventing the prototypes from collapsing and also encouraging the model to learn mono-semantic features.

Furthermore, we introduced the Top-K Mono-semanticity Scale, a new metric to quantitatively evaluate the explainability of not only prototype-based models, but also other explainable methods relying on sample-based explanations. This metric enables a systematic assessment of how well each prototype aligns with a single understandable concept. In multiple experiments, both the quantitative and qualitative results demonstrate the SPN can make accurate predictions and meanwhile provide explanations of its decision-making.

Despite the strengths of SPN, there are several limitations that require further investigation, including a more comprehensive evaluation of general explainability, a more transparent generation module, and larger scales of prototypes. On top of that, our future work will focus on: 

1) Extending SPN to a larger scale of data. So far, SPN only shows promising results with the TITAN and PIE datasets, due to the limitation of data with rich action labels. Future research should focus on extending SPN to larger and more diverse datasets, as well as exploring unsupervised learning to utilize data without manual labels.

2) Exploiting knowledge from well-trained models, such as LLMs. Due to the flexibility of natural language, there have been relatively mature evaluation metrics for the explainability of LLMs. Although it is difficult to directly apply these metrics to multi-modal tasks, it is still feasible and worth investigating to develop explainable multi-modal methods on top of a transparent and trustworthy language model, 

\addtolength{\textheight}{-3cm}   





\balance
\bibliographystyle{IEEEtran}
\bibliography{ref}

\end{document}